\documentclass{article}

\usepackage{arxiv}
\usepackage[ruled,vlined]{algorithm2e}

\usepackage[utf8]{inputenc} 
\usepackage[T1]{fontenc}    
\usepackage{hyperref}       
\usepackage{url}            
\usepackage{booktabs}       
\usepackage{amsfonts}       
\usepackage{nicefrac}       
\usepackage{microtype}      
\usepackage{lipsum}		
\usepackage{graphicx}
\usepackage{natbib}
\usepackage{doi}
\usepackage{mathtools}

\DeclarePairedDelimiter\floor{\lfloor}{\rfloor}

\usepackage{placeins}

\usepackage{graphicx}
\usepackage{amsmath}
\usepackage{longtable}

\usepackage[table,xcdraw]{xcolor}

\graphicspath{ {./images/} }

\title{Improving Time Series Classification Algorithms Using Octave-Convolutional Layers}

\author{Samuel ~Harford \\
	Department of Mechanical \& Industrial Engineering\\
	University of Illinois at Chicago\\
	\texttt{sharfo2@uic.edu} \\
	\And
	Fazle ~Karim \\
	Department of Mechanical \& Industrial Engineering\\
	University of Illinois at Chicago\\
	\texttt{karim1@uic.edu }\\
	\And
	Houshang ~Darabi \\
	Department of Mechanical \& Industrial Engineering\\
	University of Illinois at Chicago\\
	\texttt{hdarabi@uic.edu}

}

\date{}

\renewcommand{\shorttitle}{\textit{arXiv} Template}

\hypersetup{
pdftitle={A template for the arxiv style},
pdfsubject={q-bio.NC, q-bio.QM},
pdfauthor={David S.~Hippocampus, Elias D.~Striatum},
pdfkeywords={First keyword, Second keyword, More},
}

\begin{document}
\maketitle

\begin{abstract}
Deep learning models utilizing convolution layers have achieved state-of-the-art performance on univariate time series classification tasks. In this work, we propose improving CNN based time series classifiers by utilizing Octave Convolutions (OctConv) to outperform themselves. These network architectures include Fully Convolutional Networks (FCN), Residual Neural Networks (ResNets), LSTM-Fully Convolutional Networks (LSTM-FCN), and Attention LSTM-Fully Convolutional Networks (ALSTM-FCN). The proposed layers significantly improve each of these models with minimally increased network parameters. In this paper, we experimentally show that by substituting convolutions with OctConv, we significantly improve accuracy for time series classification tasks for most of the benchmark datasets. In addition, the updated ALSTM-OctFCN performs statistically the same as the top two time series classifers, TS-CHIEF and HIVE-COTE (both ensemble models). To further explore the impact of the OctConv layers, we perform ablation tests of the augmented model compared to their base model. 
\end{abstract}


\section{Introduction}

Time series data is ubiquitous. It exists in weather readings \cite{taylor2009wind}, financial recordings \cite{tsay2005analysis}, industrial observations \cite{alwan1988time}, and psychological signals \cite{jebb2015time, kadous2002temporal}. Over the past two decades, classifying these time series data has received enormous attention, and their models are constantly improving. The main objectives of these time series classification algorithms are to perform accurately and efficiently \cite{sharabiani2018asymptotic}.

Several models have been proposed to classify time series data accurately. Typically, these models are feature-based, ensembles, or deep learning models. Some promising feature-based univariate time series classification models are Bag-of-Words (BoW) \cite{lin2007experiencing}, Bag-of-features (TSBF) \cite{baydogan2013bag}, Bag-of-SFA-Symbols (BOSS) \cite{schafer2015boss}, BOSSVS \cite{schafer2016scalable}, and Word ExtrAction for time Series cLassification (WEASEL) \cite{schafer2017fast}. In 2007, Lin et al. propose representing time series data using a symbolic representation, which is fed into a BoW classifier \cite{lin2007experiencing}. Subsequently, Baydogan et al. propose utilizing multiple subsequences of random lengths to capture local information of time series data (TSBF) \cite{baydogan2013bag}. These subsequences act as a cookbook to classify time series data. In 2015 and 2016, Schafer et al. proposed BOSS \cite{schafer2015boss} and BOSSVS \cite{schafer2016scalable} representing time series data symbolically utilizing histograms and a symbolic Fourier approximation. BOSSVS is an extension of BOSS that reduces the time complexity by utilizing a vector space model. In 2017, Schafer and Leser proposed WEASEL \cite{schafer2017fast}, an accurate and scalable time series classification model. WEASEL extracts important features by applying a chi-square test on discriminative words that are obtained through a symbolic Fourier approximation. Fast logistic regression utilizes these features to classify time series data. Around the same time, Flynn et al. present RISE, which extracts features of a time series in the time domain (using autocorrelation function, partial autocorrelation function, and an autoregressive model) and features from the frequency domain using power spectrum \cite{flynn2019contract}. A random forest is trained on these extracted features to classify the time series data. Nguyen et al. propose utilizing multiple resolutions of symbolic representations, multiple domains representations, and efficient navigation of a huge symbolic word space via a modified symbolic sequence classifier to achieve state-of-the-art results.

Several ensemble models, such as Elastic Ensemble (EE) \cite{lines2015time}, Shapelet Ensembles (SE) \cite{bagnall2015time}, Flat Collective of Transformation-Based Ensembles (FLAT-COTE) \cite{bagnall2015time}, Hierarchical Vote COTE (HIVE-COTE) \cite{lines2016hive}, and Time Series Combination of Heterogeneous and Integrated Embeddings \cite{shifaz2020ts} and have yielded state-of-the-art performance. EE integrates 11 1-NN algorithms via a weighted ensemble method. SE integrates a heterogeneous ensemble on a transformed shapelet. FLAT-COTE. FLAT-COTE combines 35 various time series classifiers into a single classifier by applying a weighted majority vote. HIVE-COTE extends FLAT-COTE by integrating the BOSS and RISE algorithms. More recently, TS-CHIEF \cite{shifaz2020ts} utilizes tree-structured classifiers to efficiently ensemble classifiers yielding results similar to HIVE-COTE for a fraction of the time. 

Deep learning models have received much attention. Fully Convolutional Network (FCN) \cite{wang2017time}, Residual Neural Network (ResNet) \cite{wang2017time}, LSTM-FCN, and ALSTM-FCN are currently the state-of-the-art deep learning algorithms for time series classification \cite{karim2017lstm}. All these algorithms take advantage of convolutional neural network (CNN) layers to extract features to classify univariate time series data. This paper proposes improving all CNN based time series classifiers by substituting the CNN layers with a modified Octave-Convolutional layer. An OctConv layer factorizes the mixed feature maps of a convolutional layer by their frequencies. OctConv has never been applied to time series data or on 1D vectors. Results indicate OctConv to statistically improve the classification accuracy of the various time series deep learning architectures.  

The remainder of this paper is structured as follows: Section \ref{section:2} provides background on univariate time series classifiers and neural network architectures. Section \ref{section:3} details the proposed methods for this work. Section \ref{section:4} presents the experiments and results of our proposed methodology on the benchmark univariate time series classifiers. Section \ref{section:5} concludes the paper and discusses future work.

\section{Literature Review}
\label{section:2}
\subsection{Fully Convolutional Network}
\label{sec:sub:fcn}
FCNs have proven to be effective deep learning classifiers in many domains including computer vision \cite{wang2015visual} and natural language processing \cite{tachibana2018efficiently}. Wang et al. introduced the use of FCNs as a good base for a deep learning based time series classifier \cite{wang2017time}. The input block accepts a univariate time series. This is then passed to three convolutional blocks where each block has a 1-D convolutional layer, followed by batch normalization, and finally a ReLU activation. The convolutional layers have filter sizes of \{128, 256, 128\} and kernel sizes of \{8, 5, 3\}, respectively. The convolutional blocks are followed by a global average pooling layer and finally a softmax layer to output a class probability vector. 


\subsection{Residual Neural Network}
\label{sec:sub:res}
ResNets utilize deeper structures of networks and employ skip connections to utilize information from bottom level gradients. ResNets are often utilized in object detection \cite{jung2017resnet}. Wang et al. explore the use of ResNets for time series classification \cite{wang2017time}. The input block accepts a univariate time series. This is then passed to three residual blocks. Each residual block goes through a 1-D convolutional layer, batch normalization, and ReLU layer three times. The final ReLU layer of each residual block has a skip connection to the input of the block. The filter size for convolutional layers in the same block is the same where the sizes are \{64, 128, 128\} for the respective residual blocks. The three residual blocks are then followed by a global average pooling layer and a softmax output. 


\subsection{LSTM-FCN \& ALSTM-FCN}
\label{sec:sub:lfcn}
The LSTM-FCN architecture aims to extract feature information by branching the input \cite{karim2017lstm}. The input block accepts a univariate time series. The input is then fed to two branches. One branch is an FCN with the same parameters defined in Section \ref{sec:sub:fcn}. The other branch performs a dimension shuffle on the input. This step transposes the input shape from a univariate time series to a multivariate time series where each channel is a single time step. Without the dimension shuffle the model training results in severe overfitting. This branch then passes to an LSTM layer or Attention LSTM layer \cite{bahdanau2016end}. The final step of this branch is a dropout layer. The result of the two branches is then concatenated and fed to a softmax output layer. 

\subsection{Octave Convolution Layer}
The Octave Convolution is formulated as a replacement for general convolutional layers that reduces the spatial redundancy and computational cost \cite{chen2019drop}. An OctConv operation consists of decomposing the input into a high and low frequency. Let the high and low frequency output of the OctConv be represented by $Y = \{Y^H, Y^L\}$ where the frequency components are calculated as:
\begin{align*}
Y^H= Y^{H \rightarrow H}+ Y^{L \rightarrow H}\\
Y^L= Y^{L \rightarrow L}+ Y^{H \rightarrow L}
\end{align*}
where $Y^{H \rightarrow H}$ and $Y^{L \rightarrow L}$ represent the intra-frequency updates, and $Y^{L \rightarrow H}$ and $Y^{H \rightarrow L}$ are the inter-frequency communications. The intra-frequency component is used to update the information within each frequency, while inter-frequency communication further enables information exchange between frequencies.

\begin{figure}[ht]
\centering
\includegraphics[width=9cm]{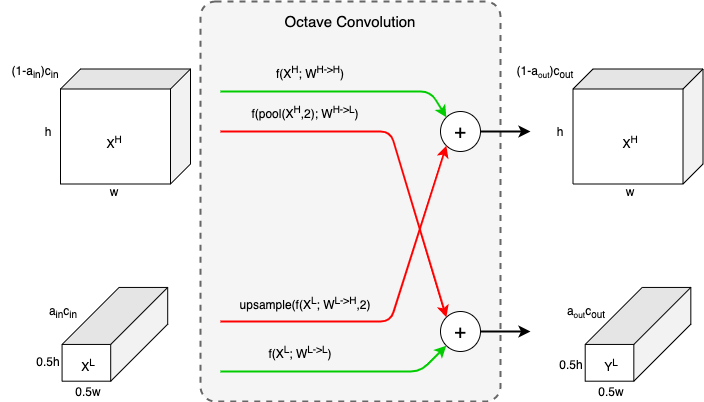}
\caption{Octave Convolution.}
\label{oct_conv}
\end{figure}

These intra and inter frequencies can are calculated using the convolutional kernel $W$ and input tensor $X$. The representation of $W$ is separated into high and low components $W = \{W^H, W^L\}$ responsible for convolving with the respective input component $X= \{X^H, X^L\}$. Each of these components can further be divided into the intra and inter frequencies $W^H = \{W^{H \rightarrow H}, W^{H \rightarrow L}\}$ and $W^L = \{W^{L \rightarrow H}, W^{L \rightarrow L}\}$. This breakdown of components formalize the output tensor calculations as:

\begin{align*}
Y_{p,q}^H = Y^{H \rightarrow H}+ Y^{L \rightarrow H}\\
= \sum_{i,j \in \mathcal{N}_k} W^{H \rightarrow H}_{i+\tfrac{k-1}{2},j+\tfrac{k-1}{2}} \top X^{H}_{p+i, q+i}\\
+ \sum_{i,j \in \mathcal{N}_k} W^{L \rightarrow H}_{i+\tfrac{k-1}{2},j+\tfrac{k-1}{2}} \top X^{L}_{\floor{\frac{p}{2}}+i, \floor{\frac{q}{2}}+i}
\end{align*}

\begin{align*}
Y_{p,q}^L = Y^{L \rightarrow L}+ Y^{H \rightarrow L}\\
= \sum_{i,j \in \mathcal{N}_k} W^{L \rightarrow L}_{i+\tfrac{k-1}{2},j+\tfrac{k-1}{2}} \top X^{L}_{p+i, q+i}\\
+ \sum_{i,j \in \mathcal{N}_k} W^{H \rightarrow L}_{i+\tfrac{k-1}{2},j+\tfrac{k-1}{2}} \top X^{H}_{2*p+0.5+i, 2*q+0.5+i}
\end{align*}
where $(p,q)$ is a location coordinate, $\mathcal{N}_k = \{(i, j):i =\{\frac{-k-1}{2}, \dots , \frac{k-1}{2}\}, j =\{\frac{-k-1}{2}, \dots , \frac{k-1}{2}\}\}$ is a neighboring coordinate, and $\floor{}$ represents the floor operation. Figure \ref{oct_conv} illustrates the interactions of the Octave Convolution. 

\section{Methodology}
\label{section:3}
\subsection{1D Octave Convolutional Layer}
\label{section:3a}
Deep learning models have proven to be successful models to classify univariate time series \cite{wang2017time}. The majority of deep learning models for time series classification utilize a series of CNN layers as the primary feature extractor of the neural network architecture. While CNNs have proven to be effective for a variety of deep learning classification tasks, there remains room for improvement in the feature extraction of these layers. In this work, we propose extensions of deep learning models for time series classification with the use of OctConv layers instead of CNN layers. We replace the convolutional layers for Octave Convolutions, OctConv Fully Convolutional Network (OctFCN), OctConv Residual Neural Network (OctResNet), LSTM-OctConv Fully Convolutional Network (LSTM-OctFCN), Attention LSTM-OctConv Fully Convolutional Network and (ALSTM-OctFCN) we use are for the OctConv layer to the case of 1D sequence models to enhance classification accuracies for each model.

\begin{figure*}[]
\centering
\includegraphics[width=17cm]{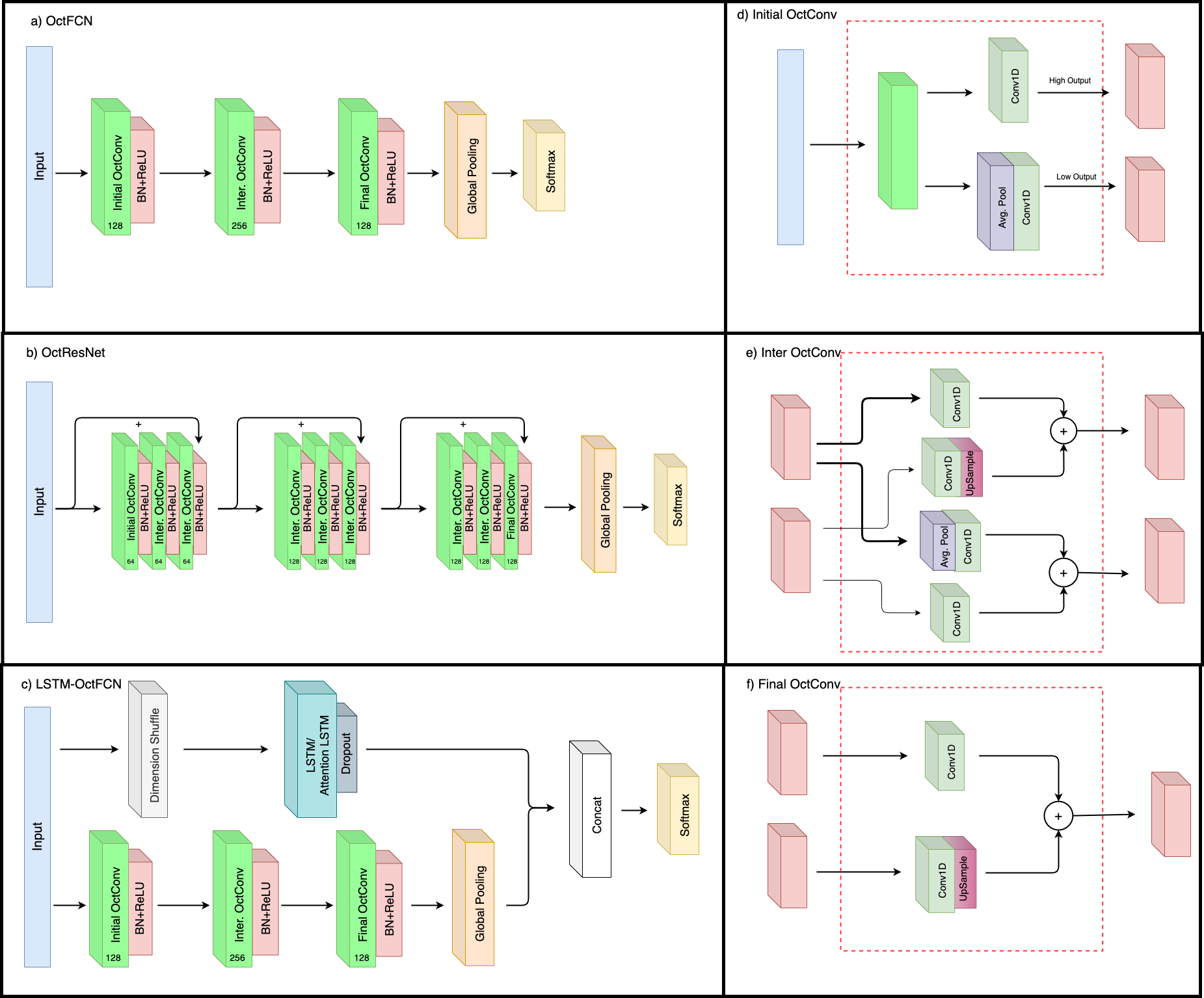}
\caption{Neural Network Augmentations and OctConv Blocks}
\label{method}
\end{figure*}

The OctConv augmentation to the models in Figure \ref{method}D-F aims to enhance the feature extraction at each layer. The initial OctConv layer takes the same input shape as the original convolutional layer. The filter size $f_{init}$ for the initial layer is split based on a parameter $\alpha$ which is set to 0.5. The high filter is achieved by feeding the input to a 1-D convolutional layer with filter size of $\alpha * f_{init}$. The low filter is achieved by first downsampling the input through a 1D average pool and then fed to a 1-D convolutional layer with filter size of $(1-\alpha) * f_{init}$. The initial layer then outputs a high and low filter component. The intermediate layers of the OctConv blocks take as input a high and low component and again split each component by the factor $\alpha$ and the filter size $f_{inte}$. The high$\rightarrow$high component is obtained through a 1-D convolution of the high input with a filter size of $(1-\alpha) * f_{inte}$. The low$\rightarrow$high component is obtained through a 1-D convolution of the low input with a filter size of $(1-\alpha) * f_{inte}$ and then performing a 1-D upsampling. The final high component is calculated by an elementwise addition of the high$\rightarrow$high and low$\rightarrow$high components. The high$\rightarrow$low component is obtained through a 1-D average pooling of the high input and then fed to a 1-D convolution with a filter size of $\alpha * f_{inte}$. The low$\rightarrow$low component is obtained through a 1-D convolution of the low input with a filter size of $\alpha * f_{inte}$. The final low component is the elementwise addition of the high$\rightarrow$low and low$\rightarrow$low components. The final layer of the OctConv block takes as input a high and low component and the filter size $f_{fin}$. Both the high and low inputs are fed to 1-D convolutions with filter size $f_{fin}$. The low component is then upsampled to match shapes. The high and low components are finally added together to can the final OctConv output. 

OctConv layers aim to save on computational processing and memory through the use of the high and low filter \cite{chen2019drop}. In addition, the processing for features in the high and low filters results in better model performance overall. In this work, we aim to understand the impact of OctConv's ability to capture global information through the task of time series classification. The Octave Convolution function aims to improve performance over transitional convolutions by exploiting the low- and high-frequency information and increasing the field size to learn better global information.

\subsection{Network Architectures}
\label{section:3b}
In this section, we describe the model architectures for the four most common CNN based time series classifiers. 

\subsubsection{OctConv Fully Convolutional Network}
The OctFCN architecture is an extension of the FCN introduced in Section \ref{sec:sub:fcn}. Figure \ref{method}A illustrates the OctFCN architecture. The input block accepts a univariate time series. This is then passed to three OctConv blocks where each block has a 1-D Octave Convolutional layer, followed by batch normalization, and finally a ReLU activation. The 1-D Octave Convolutional layers have filter sizes of \{128, 256, 128\} and kernel sizes of \{8, 5, 3\}, respectively. The three OctConv blocks are followed by a global average pooling layer and finally a softmax layer to output a class probability vector. 

\subsubsection{OctConv Residual Neural Network}
The OctResNet architecture is an extension of the ResNet introduced in Section \ref{sec:sub:res}. Figure \ref{method}B illustrates the OctResNet architecture. The input block accepts a univariate time series. This is then passed to three modified residual blocks. Each of these blocks goes through a 1-D Octave Convolutional layer, batch normalization, and ReLU layer three times. The final ReLU layer of each residual block has a skip connection to the input of the block. The filter size for 1-D Octave Convolutional layers in the same block is the same where the sizes are \{64, 128, 128\} for the respective blocks. The three modified residual blocks are then followed by a global average pooling layer and a softmax output. 

\subsubsection{A/LSTM-OctConv Fully Convolutional Network}
The LSTM-OctFCN architecture is an extension of the LSTM-FCN introduced in Section \ref{sec:sub:lfcn}. Figure \ref{method}C illustrates the LSTM-OctFCN architecture. The input block accepts a univariate time series. The input is then fed to two branches. One branch is an OctFCN with three OctConv blocks where each block has a 1-D Octave Convolutional layer, followed by batch normalization, and finally a ReLU activation. The 1-D Octave Convolutional layers have filter sizes of \{128, 256, 128\} and kernel sizes of \{8, 5, 3\}, respectively. The other branch performs a dimension shuffle on the input. This step transposes the input shape from a univariate time series to a multivariate time series where each channel is a single time step. The final step of this branch is a dropout layer. This branch is not modified because there are no CNN layers. The result of the two branches is then concatenated and fed to a softmax output layer. 

\subsection{Training \& Evaluation Methodology}
\label{t_e_methods}
Time series datasets are defined as a tensor of shape (N, Q, M), where N is the number of samples, Q is the maximum length of a time series, and M is the number of dimensions. In the case of univariate time series M is 1 for all datasets. While this study focuses on univariate time series classification, the models can easily be extended to multivariate applications. 

The primary modeling parameters for each network architecture are defined in Section \ref{section:3a}. The majority of these parameters are constant for all datasets that are modeling. During the training of the neural networks and their OctConv variants, a gridsearch of optimal parameters is not conducted. The objective of these experiments is to compare the augmentation of the convolution layers, not to obtain the optimal model. The only exception is the number of units in the LSTM for the LSTM-FCN and LSTM-OctFCN models. These units have been optimized in previous works and are utilized in our study \cite{karim2017lstm, karim2019insights}. 

In this paper, various models are evaluated using accuracy and rank. To fairly compare these models, we use an average of 20 runs for each model.

\section{Experiments \& Results}
\label{section:4}
\subsection{Univariate Time Series Benchmarks}
The current University of California-Riverside (UCR) time series classification benchmark consists of 128 univariate datasets \cite{dau2019ucr}. The first 85 datasets were introduced in the 2015 version of the benchmarks. These datasets consist of univariate time series with uniform length and z-normalization preprocessing. The additional 43 datasets in the current version aim to address the issues of the data and task the user with more realistic classification benchmarks. This includes the addition of varying length time series and removes the preprocessing of time series benchmarks. 

\subsection{VS Deep Learning Classifiers}
This study aims to show the impact of the OctConv layer as a replacement for a traditional convolutional layer. This replacement is evaluated by comparing neural networks trained using the convolutional layer compared to the same architecture with OctConv in place of the convolutional layers. These neural networks include the FCN, ResNet and LSTM-FCN introduced in Figure \ref{method}. In addition, a modified LSTM-FCN for an attention lstm (ALSTM-FCN) is evaluated during the experiments.  

To compare a neural network to its OctConv version, the average accuracy of 20 trained networks is used as the final metric for each classifier. The Wilcoxon signed-rank test (WSRT) is a non-parametric statistical test that hypothesizes the median of the rank between the compared models is similar. The alternative hypothesis of the WSRT is that the median of the rank is not similar for compared models. The WSRT is used to evaluate the performance of base neural networks with their OctConv versions. Table \ref{tab:dl_wsrt} shows the results of the WSRT for the four neural networks explored in this study. At a p-value of 0.05 all of the evaluated networks show a significant improvement in accuracy across the benchmark datasets. This shows the impact of the OctConv in place of the traditional convolution results in significantly better models in a variety of neural networks for the task of time series classification.
\begin{table}[h]
\center
\caption{Wilcoxon Signed-Rank Test comparing AVERAGE testing accuracy of Models developed with traditional convolutions vs OctConvs (p-value shown)}
\begin{tabular}{|c|c|}
\hline
Model     & Oct P-Value                      \\ \hline
FCN       & \cellcolor[HTML]{34FF34}4.60E-07 \\ \hline
Resnet    & \cellcolor[HTML]{34FF34}3.19E-13 \\ \hline
LSTM-FCN  & \cellcolor[HTML]{34FF34}3.51E-10 \\ \hline
ALSTM-FCN & \cellcolor[HTML]{34FF34}1.63E-09 \\ \hline
\end{tabular}
\label{tab:dl_wsrt}

\end{table}

In addition to the average accuracy, the maximum accuracy of 20 trained networks is compared. Table \ref{tab:dl_wsrt_max} shows the results of the WSRT for the four neural networks based on the maximum accuracy. At a p-value of 0.05 all networks except ALSTM-FCN show to have a significant improvement in maximum accuracy on the benchmark datasets. 
\begin{table}[h]
\center
\caption{Wilcoxon Signed-Rank Test comparing MAXIMUM testing accuracy of Models developed with traditional convolutions vs OctConvs  (p-value shown)}
\begin{tabular}{|c|c|}
\hline
Model     & Oct P-Value                      \\ \hline
FCN       & \cellcolor[HTML]{34FF34}7.44E-05 \\ \hline
Resnet    & \cellcolor[HTML]{34FF34}1.33E-03 \\ \hline
LSTM-FCN  & \cellcolor[HTML]{34FF34}9.21E-04 \\ \hline
ALSTM-FCN & \cellcolor[HTML]{FE0000}5.22E-02 \\ \hline
\end{tabular}
\label{tab:dl_wsrt_max}

\end{table}

\begin{figure}[ht]
\centering
\includegraphics[width=9cm]{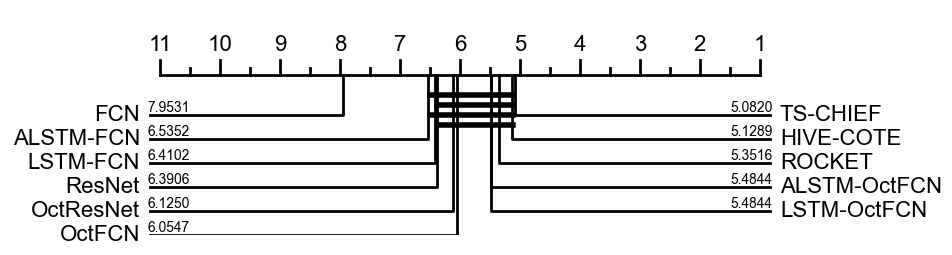}
\caption{Critical difference diagram of the arithmetic means of the ranks for Deep Learning Models}
\label{cd_diagram}
\end{figure}

Figure \ref{cd_diagram} illustrates a critical difference diagram for the models evaluated in this study. The critical difference diagram is conducted using a Wilcoxon-Holm post-hoc analysis to evaluate the arithmetic means of the ranks \cite{fawaz2019deep}. This figure confirms that each neural network has a lower rank than that of its original version.

\begin{figure*}[ht]
\centering
\includegraphics[width=15cm]{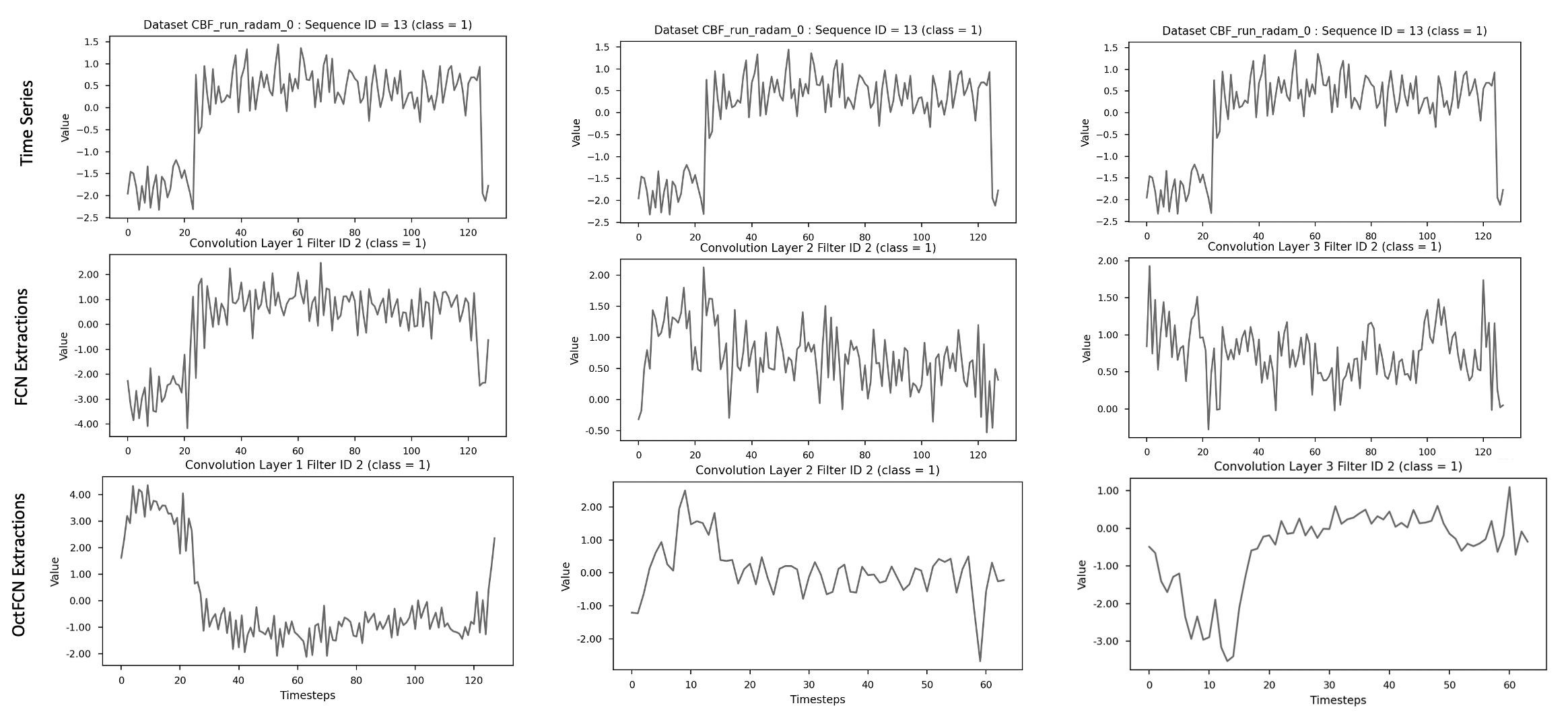}
\caption{Ablation visual representation of the input signal after transformation in FCN and OctConv through randomly selected filters from each layer}
\label{ablation}
\end{figure*}

\begin{table}[h]
\center
\caption{Wilcoxon Signed-Rank Test comparing AVERAGE testing accuracy of Models developed with OctConvs to some of the SOTA models (p-value shown)}
\begin{tabular}{|l|c|c|c|}
\hline
Model          & \multicolumn{1}{l|}{HIVE-COTE}                          & \multicolumn{1}{l|}{TS-CHIEF}                           & \multicolumn{1}{l|}{ROCKET}                             \\ \hline
OctFCN      & \cellcolor[HTML]{34FF34}{\color[HTML]{333333} 4.15E-02} & \cellcolor[HTML]{FE0000}{\color[HTML]{333333} 9.78E-02} & \cellcolor[HTML]{34FF34}{\color[HTML]{333333} 1.06E-02} \\ \hline
OctResnet     & \cellcolor[HTML]{34FF34}{\color[HTML]{333333} 2.56E-03} & \cellcolor[HTML]{34FF34}{\color[HTML]{333333} 1.96E-02} & \cellcolor[HTML]{34FF34}{\color[HTML]{333333} 7.37E-04} \\ \hline
LSTM-OctFCN  & \cellcolor[HTML]{34FF34}{\color[HTML]{333333} 6.31E-04} & \cellcolor[HTML]{34FF34}{\color[HTML]{333333} 4.65E-03} & \cellcolor[HTML]{34FF34}{\color[HTML]{333333} 9.81E-05} \\ \hline
ALSTM-OctFCN & \cellcolor[HTML]{FE0000}7.20E-02                        & \cellcolor[HTML]{FE0000}1.12E-01                        & \cellcolor[HTML]{34FF34}{\color[HTML]{333333} 1.33E-02} \\ \hline
\end{tabular}
\label{tab:sota_wsrt}
\end{table}

\begin{figure*}[ht]
\centering
\includegraphics[width=15cm]{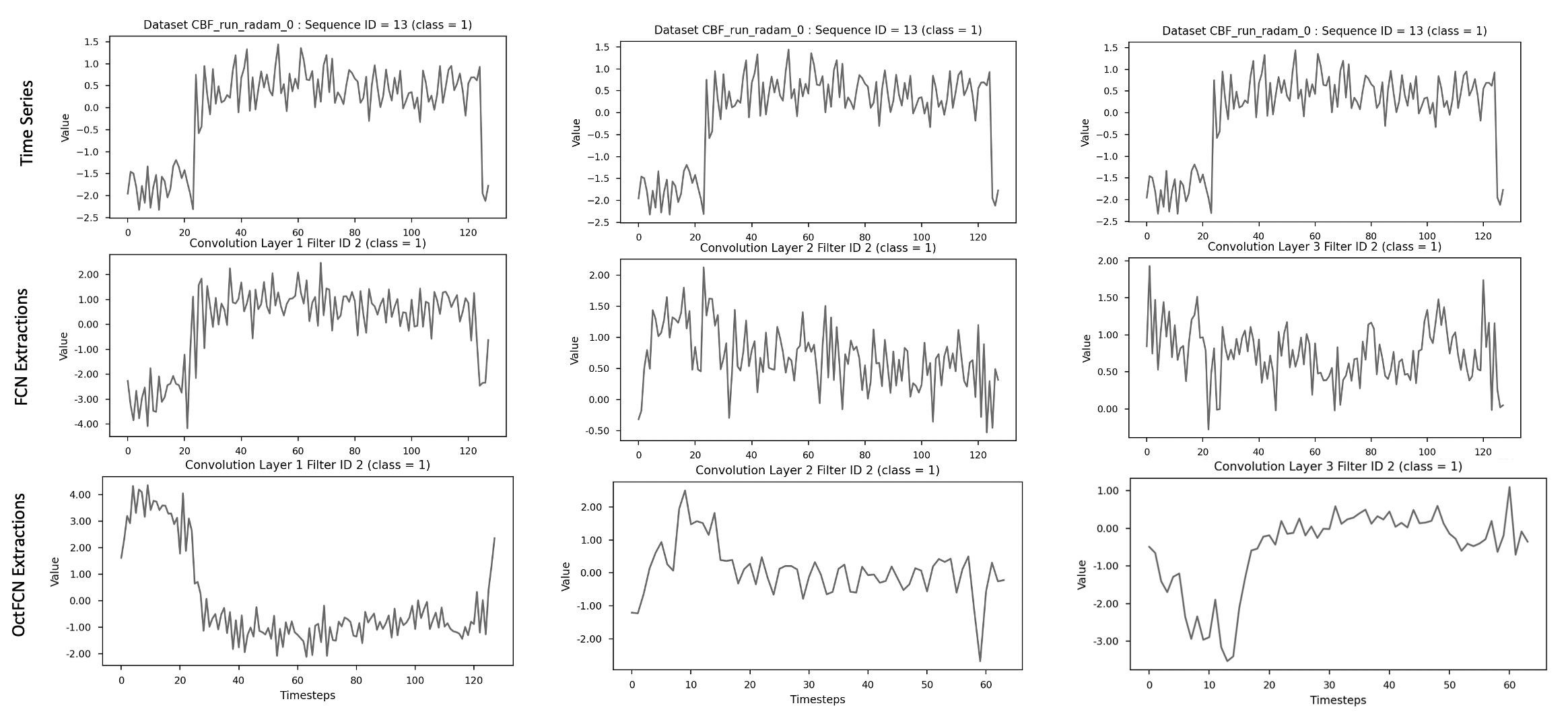}
\caption{Ablation visual representation of the input signal after transformation in FCN and OctConv through randomly selected filters from each layer}
\label{ablation}
\end{figure*}

\subsection{VS State-of-the-Art Classifiers}
As an additional evaluation, we compare our results to those reported as existing non-deep learning state-of-the-art (SOTA) models. These include models such as TS-CHIEF \cite{shifaz2020ts}, ROCKET \cite{dempster2020rocket}, HIVE-COTE \cite{lines2018time} and more. As stated in Section \ref{t_e_methods}, our model accuracies are evaluated on an average of 20 runs. For existing models, we utilize the metrics reported in the literature.

Table \ref{results_long} shows the results for the models developed in this study compared to the existing SOTA models for time series classification on the 128 benchmark datasets. The first 85 datasets are from the original achieve. These datasets have been available for research and compared more rigorously than the new datasets. For this reason, the SOTA models are more easily available for comparison through the existing literature. The remainder shows the SOTA for the remaining 43  benchmark datasets. Bold cells in these tables indicate the developed models that achieve higher accuracy than the existing SOTA. 

Figure \ref{cd_diagram} illustrates the critical difference diagram of our models and the top SOTA models.  These SOTA models often rely on ensemble techniques to boost optimal performance. The purpose of this study is not to statistically outperform SOTA models on the benchmark, rather to illustrate the effectiveness of the OctConv layer in the place of convolutional layers. Table \ref{tab:sota_wsrt} shows the results of the WSRT for the four OctConv neural networks compared to TS-CHIEF, ROCKET, and HIVE-COTE. At a significance level of 0.05, both OctResNet and LSTM-OctFCN is outperformed by HIVE-COTE, TS-CHIEF, and ROCKET. Oct-FCN performs statistically the same as TS-CHIEF. Interestingly, ALSTM-FCN performs statistically the same as the top 2 time series classifiers, HIVE-COTE and TS-CHIEF. This result shows that a simple update of a convolution layer to an octave convolution layer, not only improves its own performance, but leads ALSTM-OctFCN to perform similarly to top two SOTA TSC classifiers (HIVE-COTE and TS-CHIEF).

\subsection{Ablation}
As an additional evaluation for the impact of the OctConv layer, an ablation study is conducted to show the impact of the replacement for feature extraction. Figure \ref{ablation} shows the differences in input transformation of FCN and OctFCN for a single input of the CBF dataset. These visuals depict randomly selected filters for each of the models. Both models show clear learned feature extraction in the series of layers, however the two models result in very different noise reduction steps. While these visuals clearly show the superior feature extraction of the OctConv layers, we further study the outcomes through the use of a linear kernel classifier.

In order to explicitly determine the impact of these different methods, we evaluate the features from the developed models when applied to an SVM classifier with a linear kernel. For each of the developed models (FCN, ResNet, LSTM-FCN, and ALSTM-FCN) the features are fed to SVM classifiers to get an explicit evaluation of the extracted features. For the LSTM-FCN and ALSTM-FCN models, only the FCN component of the network is analyzed in this ablation study.

Table \ref{tab:ab_wsrt} shows the results of the WSRT comparing the accuracy for an SVM developed with convolutional features vs an SVM developed with OctConv features. These results show that (at a p-value of 0.05) there is a significant improvement of the SVM developed with OctConv features than that of the convolutional features. Based on this analysis, we can conclude that the OctConv is doing a better job creating linearly separable classes than the convolutional layer. We postulate that the OctConv layers are better because the high and low filters do a better job of capturing the global information at each layer. 

\begin{table}[h]
\center
\caption{Wilcoxon Signed-Rank Test comparing the SVM models developed from traditional convolutions vs OctConvs information (p-value shown)}
\begin{tabular}{|c|c|}
\hline
Model     & P-Value                          \\ \hline
FCN       & \cellcolor[HTML]{34FF34}6.06E-05 \\ \hline
Resnet    & \cellcolor[HTML]{34FF34}8.67E-08 \\ \hline
LSTM-FCN  & \cellcolor[HTML]{34FF34}1.70E-08 \\ \hline
ALSTM-FCN & \cellcolor[HTML]{34FF34}8.35E-06 \\ \hline
\end{tabular}
\label{tab:ab_wsrt}

\end{table}

\section{Conclusion}
\label{section:5}
In this work, we propose the replacement of traditional convolutional layers with Octave Convolutional layers. We evaluate this replacement on four state-of-the-art neural network architectures including FCN, ResNet, LSTM-FCN, and ALSTM-FCN. The experiments for this study utilize the 128 datasets provided by the University of California-Riverside time series classification archive. These experiments show that on an average of 20 runs on each dataset the tested neural networks perform significantly better when the convolutional layers are replaced with Octave Convolutional layers. Based on the average testing accuracy of 20 runs, ALSTM-OctFCN performs statistically the same as the top two time series classifiers, TS-CHIEF and HIVE-COTE. To further evaluate the replaced layers, we provide an ablation study of the features extracted by each neural network classifier. This ablation study establishes that the features extracted by the Octave Convolutional layers provide better information than that of the traditional convolutional layers. Future research in this area can explore the use of other alternatives to convolutional layers that may outperform the Octave Convolution. In addition, the generated models may be considered in an ensemble model to achieve state-of-the-art performance on the time series classification benchmarks. 

\FloatBarrier

\begin{longtable}{|c|c|c|c|c|c|c|c|c|c|}

\caption{ Classification Accuracy for the Benchmark Datasets}

\\
\textbf{Dataset} & \textbf{FCN} & \textbf{Oct} & \textbf{ResNet} & \textbf{Oct} & \textbf{LSTM} & \textbf{LSTM} & \textbf{ALSTM} & \textbf{ALSTM} & \textbf{SOTA} \\ 
\tiny{.}               & \tiny{.} & \textbf{FCN} & \tiny{.} & \textbf{ResNet} & \textbf{-FCN} & \textbf{-OctFCN} & \textbf{-FCN} & \textbf{-OctFCN} & \tiny{.} \\ \hline
Adiac                          & 0.768        & 0.801           & 0.833           & 0.828              & 0.826             & 0.823                & 0.811              & 0.821                 & \textbf{0.857}         \\ \hline
ArrowHead                      & 0.816        & 0.855           & 0.813           & 0.846              & 0.869             & \textbf{0.884}                & 0.813              & 0.868                 & 0.880          \\ \hline
ChlorineConcentration          & 0.800        & 0.800           & 0.817           & 0.813              & 0.800             & 0.811                & 0.798              & 0.807                 & \textbf{0.875}         \\ \hline
InsectWingbeatSound            & 0.382        & 0.451           & 0.448           & 0.541              & 0.649             & \textbf{0.654}                & 0.385              & 0.454                 & 0.653         \\ \hline
Lighting7                      & 0.740        & 0.814           & 0.703           & 0.784              & 0.745             & 0.775                & 0.749              & 0.814                 & \textbf{0.863}         \\ \hline
Wine                           & 0.596        & 0.778           & 0.759           & 0.681              & 0.672             & 0.770                & 0.717              & 0.756                 & \textbf{0.889}        \\ \hline
WordsSynonyms                  & 0.532        & 0.607           & 0.609           & 0.655              & 0.657             & 0.668                & 0.544              & 0.608                 & \textbf{0.779}         \\ \hline
50words                        & 0.656        & 0.694           & 0.683           & 0.722              & 0.781             & 0.801                & 0.659              & 0.696                 & \textbf{0.846 }        \\ \hline
Beef                           & 0.747        & 0.680           & 0.773           & 0.757              & 0.803             & 0.757                & 0.763              & 0.720                 & \textbf{0.933}         \\ \hline
DistPhalOutAgeGroup   & 0.799        & 0.799           & 0.805           & 0.785              & 0.823             & 0.807                & 0.825              & 0.816                 & \textbf{0.835}         \\ \hline
DistPhalOutCorrect    & 0.803        & 0.800           & 0.798           & 0.790              & 0.803             & 0.798                & 0.801              & 0.794                 & \textbf{0.820}          \\ \hline
DistalPhalanxTW                & 0.776        & 0.764           & 0.766           & 0.762              & 0.779             & 0.767                & 0.783              & 0.771                 & \textbf{0.790}          \\ \hline
ECG200                         & 0.876        & 0.900           & 0.892           & 0.907              & 0.884             & 0.899                & 0.891              & 0.898                 & \textbf{0.920}          \\ \hline
ECGFiveDays                    & 0.989        & 0.993           & 0.980           & 0.991              & 0.985             & 0.990                & 0.986              & 0.993                 & \textbf{1.000}          \\ \hline
BeetleFly                      & 0.810        & 0.855           & 0.850           & 0.850              & 0.860             & 0.900                & 0.785              & 0.850                 & \textbf{0.950}          \\ \hline
BirdChicken                    & 0.935        & 0.915           & 0.925           & 0.910              & 0.900             & 0.900                & 0.935              & 0.910                 & \textbf{0.950}        \\ \hline
ItalyPowerDemand               & 0.959        & 0.956           & 0.961           & 0.958              & 0.959             & 0.956                & 0.959              & 0.956                 & \textbf{0.970 }        \\ \hline
SonyAIBORobotSur           & 0.964        & 0.957           & 0.916           & 0.928              & 0.938             & 0.934                & 0.961              & 0.961                 & \textbf{0.985}         \\ \hline
SonyAIBORobotSurII         & 0.972        & 0.965           &\textbf{ 0.980}           & 0.970              & 0.972             & 0.962                & 0.976              & 0.966                 & 0.978         \\ \hline
MidPhalOutAgeGroup   & 0.749        & 0.736           & 0.751           & 0.733              & 0.742             & 0.737                & 0.746              & 0.742                 & \textbf{0.814}         \\ \hline
MidPhalOutCor    & \textbf{0.817}        & 0.798           & 0.816           & 0.791              & 0.814             & 0.806                & 0.816              & 0.800                 & 0.808        \\ \hline
MiddlePhalanxTW                & 0.585        & 0.605           & 0.595           & 0.593              & 0.595             & 0.599                & 0.594              & 0.593                 & \textbf{0.612 }        \\ \hline
ProxPhalOutAgeGroup & 0.833        & 0.835           & 0.853           & 0.844              & 0.845             & 0.837                & 0.845              & 0.835                 & \textbf{0.883}          \\ \hline
ProxPhalOutCorrect  & 0.902        & 0.910           & 0.914           & 0.908              & 0.906             & 0.910                & 0.906              & 0.910                 & \textbf{0.918}         \\ \hline
ProximalPhalanxTW              & 0.811        & 0.802           & \textbf{0.816}           & 0.791              & 0.814             & 0.799                & 0.810              & 0.805                 & 0.815         \\ \hline
MoteStrain                     & 0.918        & 0.906           & 0.909           & 0.901              & 0.927             & 0.908                & 0.920              & 0.903                 & \textbf{0.950}         \\ \hline
MedicalImages                  & 0.768        & 0.753           & 0.756           & 0.755              & 0.768             & 0.751                & 0.771              & 0.760                 & \textbf{0.792}         \\ \hline
Strawberry                     & 0.964        & 0.965           & 0.964           & 0.962              & 0.963             & 0.965                & 0.966              & 0.966                 & \textbf{0.980}         \\ \hline
ToeSegmentation1               & \textbf{0.975}        & 0.967           & 0.965           & 0.964              & 0.971             & 0.965                & 0.970              & 0.964                 & 0.974         \\ \hline
Coffee                         & \textbf{1.000}        & \textbf{1.000 }          & \textbf{1.000}           & \textbf{1.000}              & \textbf{1.000}             & \textbf{1.000}                & \textbf{1.000}              & \textbf{1.000}                 & \textbf{1.000}         \\ \hline
Cricket\_X                     & 0.773        & 0.791           & 0.788           & 0.814              & 0.746             & 0.762                & 0.782              & 0.791                 & \textbf{0.821}         \\ \hline
Cricket\_Y                     & 0.764        & 0.798           & 0.798           & 0.819              & 0.759             & 0.798                & 0.770              & 0.795                 & \textbf{0.826}        \\ \hline
Cricket\_Z                     & 0.774        & 0.812           & 0.804           & 0.823              & 0.786             & 0.809                & 0.783              & 0.812                 & \textbf{0.836}         \\ \hline
uWaveGestLib\_X         & 0.757        & 0.773           & 0.763           & 0.782              & 0.832             & 0.834                & 0.766              & 0.774                 & \textbf{0.855}         \\ \hline
uWaveGestLib\_Y         & 0.649        & 0.674           & 0.652           & 0.682              & 0.727             & 0.745                & 0.668              & 0.676                 & \textbf{0.759}          \\ \hline
uWaveGestLib\_Z         & 0.737        & 0.745           & 0.736           & 0.747              & 0.777             & 0.773                & 0.740              & 0.745                 & \textbf{0.792}        \\ \hline
ToeSegmentation2               & 0.903        & 0.929           & 0.899           & 0.928              & 0.905             & 0.933                & 0.905              & 0.927                 & \textbf{0.962}         \\ \hline
DiatomSizeReduction            & 0.940        & 0.930           & 0.952           & 0.933              & 0.937             & 0.937                & 0.937              & 0.936                 & \textbf{0.977}       \\ \hline
Car                            & 0.913        & 0.932           & 0.888           & 0.907              & 0.917             & 0.902                & 0.915              & 0.927                 & \textbf{0.933}         \\ \hline
CBF                            & 0.994        & 0.996           & 0.995           & 0.998              & 0.994             & 0.996                & 0.993              & 0.996                 & \textbf{1.000}      \\ \hline
CinC\_ECG\_torso               & 0.832        & 0.847           & 0.809           & 0.843              & 0.890             & 0.890                & 0.830              & 0.848                 & \textbf{0.996}          \\ \hline
Computers                      & 0.844        & 0.832           & 0.814           & 0.814              & 0.775             & 0.734                & 0.836              & 0.828                 & \textbf{0.848}         \\ \hline
Earthquakes                    & 0.763        & 0.793           & 0.760           & 0.794              & 0.794             & \textbf{0.801}                & 0.783              & 0.797                 & \textbf{0.801}         \\ \hline
ECG5000                        & 0.939        & 0.941           & 0.940           & 0.940              & 0.942             & 0.942                & 0.940              & 0.941                 & \textbf{0.948}          \\ \hline
ElectricDevices                & 0.714        & 0.705           & 0.722           & 0.705              & 0.742             & 0.732                & 0.717              & 0.719                 & \textbf{0.799}         \\ \hline
FaceAll                        & 0.922        & 0.864           & 0.848           & 0.813              & 0.910             & 0.875                & 0.914              & 0.868                 & \textbf{0.929}          \\ \hline
FaceFour                       & 0.934        & 0.953           & 0.950           & 0.953              & 0.874             & 0.876                & 0.934              & 0.955                 & \textbf{1.000}          \\ \hline
FacesUCR                       & 0.933        & 0.952           & 0.942           & 0.946              & 0.917             & 0.932                & 0.934              & 0.951                 & \textbf{0.965}          \\ \hline
Fish                           & 0.937        & 0.969           & 0.971           & 0.983              & 0.958             & 0.967                & 0.953              & 0.972                 & \textbf{0.994}          \\ \hline
FordA                          & 0.919        & 0.924           & 0.921           & 0.928              & 0.918             & 0.931                & 0.915              & 0.924                 & \textbf{0.973}          \\ \hline
FordB                          & 0.880        & 0.902           & 0.916           & \textbf{0.918}              & 0.883             & 0.904                & 0.881              & 0.902                 & 0.917         \\ \hline
Gun\_Point                     & \textbf{1.000}        & \textbf{1.000}           & 0.997           & \textbf{1.000}              & 0.995             & 0.998                & \textbf{1.000}              & \textbf{1.000}                 & \textbf{1.000}          \\ \hline
Ham                            & 0.715        & 0.750           & 0.712           & 0.730              & 0.717             & 0.727                & 0.720              & 0.750                 & \textbf{0.781}          \\ \hline
HandOutlines                   & 0.521        & 0.755           & 0.880           & 0.871              & 0.842             & 0.875                & 0.623              & 0.806                 & \textbf{0.960}          \\ \hline
Haptics                        & 0.426        & 0.494           & 0.517           & 0.553              & 0.489             & 0.514                & 0.465              & 0.510                 & \textbf{0.551}          \\ \hline
Herring                        & 0.634        & 0.703           & 0.614           & 0.661              & 0.641             & 0.698                & 0.647              & \textbf{0.727}                 & 0.703         \\ \hline
InlineSkate                    & 0.374        & 0.423           & 0.482           & 0.432              & 0.465             & 0.446                & 0.424              & 0.427                 & \textbf{0.613}         \\ \hline
LargeKitApp         & 0.890        & 0.895           & \textbf{0.903}           & 0.897              & 0.685             & 0.744                & 0.897              & 0.897                 & 0.896          \\ \hline
Lighting2                      & 0.708        & 0.711           & 0.730           & 0.726              & 0.752             & 0.769                & 0.718              & 0.728                 & \textbf{0.885}          \\ \hline
MALLAT                         & 0.963        & 0.966           & 0.938           & 0.961              & 0.963             & 0.961                & 0.966              & 0.966                 & \textbf{0.980}          \\ \hline
Meat                           & 0.863        & 0.875           & 0.938           & 0.942              & 0.895             & 0.922                & 0.885              & 0.910                 & \textbf{1.000}          \\ \hline
NonInvFatECGThor1   & 0.949        & 0.958           & 0.954           & 0.955              & \textbf{0.962}             & \textbf{0.962}                 & 0.955              & 0.956                 & \textbf{0.962}          \\ \hline
NonInvFatECGThor2   & 0.929        & 0.947           & 0.949           & 0.956              & 0.954             & 0.957                & 0.948              & 0.951                 & \textbf{0.969 }        \\ \hline
OliveOil                       & 0.533        & 0.720           & 0.827           & 0.847              & 0.593             & 0.770                & 0.587              & 0.767                 & \textbf{0.933}           \\ \hline
OSULeaf                        & 0.979        & 0.970           & 0.974           & 0.965              & 0.895             & 0.907                & 0.979              & 0.971                 & \textbf{0.988}          \\ \hline
PhalOutlCor       & 0.813        & 0.821           & 0.825           & 0.821              & 0.816             & 0.817                & 0.812              & 0.817                 & \textbf{0.854}          \\ \hline
Phoneme                        & 0.336        & 0.345           & \textbf{0.354}           & 0.345              & 0.252             & 0.265                & 0.340              & 0.343                 & 0.349          \\ \hline
Plane                          & \textbf{1.000}        & \textbf{1.000}           & \textbf{1.000 }          & \textbf{1.000}              & \textbf{1.000}             & \textbf{1.000}                & \textbf{1.000}              & \textbf{1.000}                 & \textbf{1.000}          \\ \hline
RefrigerationDevices           & 0.510        & 0.528           & 0.555           & 0.534              & 0.466             & 0.480                & 0.507              & 0.525                 & \textbf{0.581}          \\ \hline
ScreenType                     & 0.637        & 0.637           & 0.648           & 0.627              & 0.545             & 0.510                & 0.639              & 0.634                 & \textbf{0.707 }         \\ \hline
ShapeletSim                    & 0.956        & 0.926           & 0.948           & 0.874              & 0.966             & 0.953                & 0.927              & 0.921                 & \textbf{1.000}          \\ \hline
ShapesAll                      & 0.885        & 0.903           & 0.912           & 0.917              & 0.890             & 0.904                & 0.884              & 0.902                 & \textbf{0.930}          \\ \hline
SmallKitApp         & 0.734        & 0.759           & 0.769           & 0.735              & 0.610             & 0.599                & 0.766              & 0.757                 & \textbf{0.853}          \\ \hline
StarlightCurves                & 0.964        & 0.975           & 0.976           & 0.977              & 0.965             & 0.974                & 0.964              & 0.975                 & \textbf{0.982}          \\ \hline
SwedishLeaf                    & 0.969        & 0.966           & 0.960           & 0.960              & \textbf{0.971}             & 0.968                & 0.968              & 0.968                 & 0.966          \\ \hline
Symbols                        & 0.973        & \textbf{0.983 }          & 0.972           & 0.980              & 0.963             & 0.963                & 0.976              & 0.984                 & 0.982          \\ \hline
synthetic\_control             & 0.985        & 0.999           & 0.994           & 0.998              & 0.978             & 0.999                & 0.980              & 0.999                 & \textbf{1.000}          \\ \hline
Trace                          & 0.250        & 0.250           & 0.250           & 0.250              & 0.250             & 0.250                & 0.250              & 0.250                 & \textbf{1.000 }         \\ \hline
Two\_Patterns                  & 0.863        & 0.942           & 0.963           & 0.991              & 0.994             & 0.984                & 0.873              & 0.943                 & \textbf{1.000}         \\ \hline
TwoLeadECG                     & 0.999        & 0.998           & 0.999           & 0.996              & 0.999             & 0.998                & 0.999              & 0.998                 & \textbf{1.000}          \\ \hline
UWaveGestLibAll         & 0.822        & 0.864           & 0.833           & 0.874              & 0.955             & 0.959                & 0.824              & 0.861                 & \textbf{0.976}         \\ \hline
Wafer                          & 0.997        & 0.999           & 0.995           & 0.997              & 0.998             & 0.999                & 0.997              & 0.998                 & \textbf{1.000 }         \\ \hline
Worms                          & 0.612        & 0.650           & 0.598           & 0.626              & 0.494             & 0.534                & 0.606              & 0.651                 & \textbf{0.805}          \\ \hline
WormsTwoClass                  & 0.773        & 0.774           & 0.771           & 0.762              & 0.683             & 0.699                & 0.781              & 0.773                 & \textbf{0.831}          \\ \hline
yoga                           & 0.860        & 0.879           & 0.878           & 0.888              & 0.903             & 0.904                & 0.879              & 0.883                 & \textbf{0.918}         \\ \hline
ACSF1            & 0.927        & 0.935           & 0.917           & 0.932              & 0.903             & 0.920                & 0.913              & \textbf{0.940}                 & 0.850          \\ \hline
AllGestWiX       & 0.615        & 0.646           & 0.707           & 0.718              & 0.693             & \textbf{0.724}                & 0.718              & 0.722                 & 0.717         \\ \hline
AllGestWiY       & 0.746        & 0.778           & 0.764           & 0.773              & 0.753             & 0.781                & 0.788              & \textbf{0.856}                 & 0.730         \\ \hline
AllGestWiZ       & 0.694        & 0.725           & 0.715           & 0.720              & 0.704             & \textbf{0.758}                & 0.733              & 0.746                 & 0.651         \\ \hline
BME              & 0.947        & 0.980           & 0.953           & 0.968              & 0.952             & 0.996                & 0.833              & 0.900                 & \textbf{0.999}        \\ \hline
Chinatown        & 0.938        & 0.956           & 0.964           & 0.973              & 0.988             & 0.988                & 0.985              & \textbf{0.995}                 & 0.970         \\ \hline
Crop             & 0.764        & 0.783           & 0.789           & 0.793              & 0.764             & \textbf{0.875}                & 0.762              & 0.799                 & 0.793         \\ \hline
DodgLpDay        & 0.614        & 0.647           & 0.647           & 0.648              & 0.634             & \textbf{0.652}                & 0.486              & 0.550                 & 0.588         \\ \hline
DodgLpGm         & 0.826        & 0.833           & 0.846           & 0.856              & 0.893             & 0.907                & 0.779              & 0.823                 & \textbf{0.928}         \\ \hline
DodgLpWnd        & 0.938        & 0.967           & 0.949           & 0.949              & 0.927             & 0.933                & 0.970              & 0.973                 & \textbf{0.986}         \\ \hline
EOGHzSgn         & 0.651        & 0.656           & 0.652           & 0.663              & 0.617             & 0.632                & 0.686              & 0.680                 & \textbf{0.884}          \\ \hline
EOGVtSgn         & 0.473        & 0.490           & 0.465           & 0.472              & 0.504             & 0.529                & 0.514              & 0.518                 & \textbf{0.815}          \\ \hline
EthLevel         & 0.737        & 0.743           & 0.753           & 0.737              & 0.725             & 0.763                & 0.738              & 0.727                 & \textbf{0.875}          \\ \hline
FrzRegTr         & 0.957        & 0.991           & 0.970           & 0.965              & 0.998             & \textbf{1.000}                & \textbf{1.000}              & \textbf{1.000}                 & 0.999         \\ \hline
FRSmlTr          & 0.829        & 0.842           & 0.816           & 0.822              & 0.807             & 0.836                & 0.878              & 0.897                 & \textbf{0.995}          \\ \hline
Fungi            & 0.983        & 0.998           & 0.994           & 0.996              & 0.984             & \textbf{1.000}                & 0.995              & \textbf{1.000}                 & 0.839         \\ \hline
GestMidAirD1     & 0.736        & 0.742           & 0.719           & 0.746              & 0.703             & \textbf{0.751}                & 0.714              & \textbf{0.751}                 & 0.639         \\ \hline
GestMidAirD2     & 0.672        & 0.677           & 0.657           & 0.669              & \textbf{0.694}             & 0.681                & 0.640              & 0.628                 & 0.608         \\ \hline
GestMidAirD3     & 0.421        & 0.440           & 0.390           & 0.433              &\textbf{ 0.452}             & 0.433                & 0.444              & 0.435                 & 0.377         \\ \hline
GestPebZ1        & 0.944        & \textbf{0.973}           & 0.941           & 0.949              & 0.947             & 0.933                & 0.921              & 0.921                 & 0.826         \\ \hline
GestPebZ2        & 0.857        & 0.883           & 0.892           & 0.864              & 0.873             & \textbf{0.913}                & 0.832              & 0.818                 & 0.779         \\ \hline
GunPtAgeSp       & 0.982        & 0.973           & 0.998           & 0.995              & 0.986             & 0.988                & \textbf{1.000 }             & 0.970                 & \textbf{1.000}          \\ \hline
GunPontMVsF      & 0.962        & 0.987           & 0.976           & 0.957              & 0.980             & 0.989                & \textbf{1.000}              & \textbf{1.000}                 & \textbf{1.000}         \\ \hline
GunPointOVsY     & 0.942        & 0.949           & 0.995           & 0.957              & 0.974             & 0.983                & \textbf{1.000}              & 0.986                 & \textbf{1.000}          \\ \hline
HouseTwenty      & 0.945        & 0.955           & 0.908           & 0.961              & 0.970             & 0.962                & 0.948              & 0.970                 & \textbf{0.979}         \\ \hline
InsEPGRegTr      & \textbf{1.000}        & \textbf{1.000}           & \textbf{1.000}           & 0.995              & \textbf{1.000}             & 0.994                & \textbf{1.000}              & 0.991                 & \textbf{1.000 }         \\ \hline
InsEPGSmlTr      & 0.928        & 0.958           & \textbf{1.000}           & 0.920              & \textbf{1.000}             & 0.981                & \textbf{1.000}              & 0.976                 & \textbf{1.000}          \\ \hline
MelbPed          & 0.972        & \textbf{0.993}           & 0.983           & 0.988              & 0.973             & 0.951                & 0.976              & 0.979                 & 0.848         \\ \hline
MxShpRegTr       & 0.928        & 0.942           & 0.975           & 0.984              & 0.963             & \textbf{0.991}                & 0.968              & 0.981                 & 0.971          \\ \hline
MxShpSmlTr       & 0.918        & 0.941           & 0.928           & 0.936              & 0.922             & 0.927                & 0.924              & 0.915                 & \textbf{0.947}          \\ \hline
PickGestWiZ      & 0.853        & 0.878           & 0.864           & 0.851              & 0.897             & 0.895                & 0.899              & \textbf{0.900}                 & 0.660         \\ \hline
PigAryPress      & 0.750        & 0.760           & 0.753           & 0.740              & 0.793             & 0.824                & 0.756              & 0.826                 & \textbf{0.977}          \\ \hline
PigArtPress      & 0.964        & 0.974           & 0.974           & 0.977              & 0.985             & \textbf{1.000}                & 0.995              & \textbf{1.000}                 & 0.975          \\ \hline
PigCVP           & 0.864        & 0.869           & 0.871           & 0.937              & 0.922             & 0.941                & 0.928              & 0.956                 & \textbf{0.961}          \\ \hline
PLAID            & 0.481        & 0.489           & 0.482           & 0.494              & 0.282             & 0.483                & 0.506              & 0.515                 & \textbf{0.840}         \\ \hline
PowerCons        & 0.973        & 0.984           & 0.973           & 0.981              & \textbf{1.000}             & \textbf{1.000}                & 0.984              & \textbf{1.000}                 & 0.993          \\ \hline
Rock             & 0.718        & 0.727           & 0.676           & 0.729              & 0.764             & 0.782                & 0.795              & 0.805                 & \textbf{0.867}          \\ \hline
SgHdGendCh2      & 0.943        & 0.969           & 0.937           & 0.959              & 0.945             & \textbf{0.986}                & 0.852              & 0.879                 & 0.969         \\ \hline
SgHdMovCh2       & 0.612        & 0.626           & 0.698           & 0.630              & 0.703             & 0.714                & 0.653              & 0.716                 & \textbf{0.891}         \\ \hline
SgHdSubCh2       & 0.919        & 0.927           & 0.925           & 0.932              & 0.917             & 0.917                & 0.902              & 0.917                 & \textbf{0.951}         \\ \hline
ShkGestWiZ       & 0.961        & \textbf{0.981}           & 0.973           & 0.966              & 0.946             & 0.992                & 0.975              & 0.974                 & 0.860         \\ \hline
SmthSub          & 0.986        & 0.983           & 0.993           & 0.988              & \textbf{1.000}             & 0.988                & \textbf{1.000}              & \textbf{1.000 }                & 0.998        \\ \hline
UMB              & 0.989        & \textbf{1.000}           & \textbf{1.000}           & 0.982              & \textbf{1.000}             & 0.966                & 0.989              & \textbf{1.000}                 & 0.983         \\ \hline
\label{results_long}

\end{longtable}


\bibliographystyle{unsrtnat}
\bibliography{references}  

\begin{thebibliography}{28}
\providecommand{\natexlab}[1]{#1}
\providecommand{\url}[1]{\texttt{#1}}
\expandafter\ifx\csname urlstyle\endcsname\relax
  \providecommand{\doi}[1]{doi: #1}\else
  \providecommand{\doi}{doi: \begingroup \urlstyle{rm}\Url}\fi

\bibitem[Taylor et~al.(2009)Taylor, McSharry, and Buizza]{taylor2009wind}
James~W Taylor, Patrick~E McSharry, and Roberto Buizza.
\newblock Wind power density forecasting using ensemble predictions and time
  series models.
\newblock \emph{IEEE Transactions on Energy Conversion}, 24\penalty0
  (3):\penalty0 775--782, 2009.

\bibitem[Tsay(2005)]{tsay2005analysis}
Ruey~S Tsay.
\newblock \emph{Analysis of financial time series}, volume 543.
\newblock John wiley \& sons, 2005.

\bibitem[Alwan and Roberts(1988)]{alwan1988time}
Layth~C Alwan and Harry~V Roberts.
\newblock Time-series modeling for statistical process control.
\newblock \emph{Journal of Business \& Economic Statistics}, 6\penalty0
  (1):\penalty0 87--95, 1988.

\bibitem[Jebb et~al.(2015)Jebb, Tay, Wang, and Huang]{jebb2015time}
Andrew~T Jebb, Louis Tay, Wei Wang, and Qiming Huang.
\newblock Time series analysis for psychological research: examining and
  forecasting change.
\newblock \emph{Frontiers in psychology}, 6:\penalty0 727, 2015.

\bibitem[Kadous et~al.(2002)]{kadous2002temporal}
Mohammed~Waleed Kadous et~al.
\newblock \emph{Temporal classification: Extending the classification paradigm
  to multivariate time series}.
\newblock University of New South Wales Kensington, 2002.

\bibitem[Sharabiani et~al.(2018)Sharabiani, Darabi, Harford, Douzali, Karim,
  Johnson, and Chen]{sharabiani2018asymptotic}
Anooshiravan Sharabiani, Houshang Darabi, Samuel Harford, Elnaz Douzali, Fazle
  Karim, Hereford Johnson, and Shun Chen.
\newblock Asymptotic dynamic time warping calculation with utilizing value
  repetition.
\newblock \emph{Knowledge and Information Systems}, 57\penalty0 (2):\penalty0
  359--388, 2018.

\bibitem[Lin et~al.(2007)Lin, Keogh, Wei, and Lonardi]{lin2007experiencing}
Jessica Lin, Eamonn Keogh, Li~Wei, and Stefano Lonardi.
\newblock Experiencing sax: a novel symbolic representation of time series.
\newblock \emph{Data Mining and knowledge discovery}, 15\penalty0 (2):\penalty0
  107--144, 2007.

\bibitem[Baydogan et~al.(2013)Baydogan, Runger, and Tuv]{baydogan2013bag}
Mustafa~Gokce Baydogan, George Runger, and Eugene Tuv.
\newblock A bag-of-features framework to classify time series.
\newblock \emph{IEEE transactions on pattern analysis and machine
  intelligence}, 35\penalty0 (11):\penalty0 2796--2802, 2013.

\bibitem[Sch{\"a}fer(2015)]{schafer2015boss}
Patrick Sch{\"a}fer.
\newblock The boss is concerned with time series classification in the presence
  of noise.
\newblock \emph{Data Mining and Knowledge Discovery}, 29\penalty0 (6):\penalty0
  1505--1530, 2015.

\bibitem[Sch{\"a}fer(2016)]{schafer2016scalable}
Patrick Sch{\"a}fer.
\newblock Scalable time series classification.
\newblock \emph{Data Mining and Knowledge Discovery}, 30\penalty0 (5):\penalty0
  1273--1298, 2016.

\bibitem[Sch{\"a}fer and Leser(2017)]{schafer2017fast}
Patrick Sch{\"a}fer and Ulf Leser.
\newblock Fast and accurate time series classification with weasel.
\newblock In \emph{Proceedings of the 2017 ACM on Conference on Information and
  Knowledge Management}, pages 637--646, 2017.

\bibitem[Flynn et~al.(2019)Flynn, Large, and Bagnall]{flynn2019contract}
Michael Flynn, James Large, and Tony Bagnall.
\newblock The contract random interval spectral ensemble (c-rise): the effect
  of contracting a classifier on accuracy.
\newblock In \emph{International Conference on Hybrid Artificial Intelligence
  Systems}, pages 381--392. Springer, 2019.

\bibitem[Lines and Bagnall(2015)]{lines2015time}
Jason Lines and Anthony Bagnall.
\newblock Time series classification with ensembles of elastic distance
  measures.
\newblock \emph{Data Mining and Knowledge Discovery}, 29\penalty0 (3):\penalty0
  565--592, 2015.

\bibitem[Bagnall et~al.(2015)Bagnall, Lines, Hills, and
  Bostrom]{bagnall2015time}
Anthony Bagnall, Jason Lines, Jon Hills, and Aaron Bostrom.
\newblock Time-series classification with cote: the collective of
  transformation-based ensembles.
\newblock \emph{IEEE Transactions on Knowledge and Data Engineering},
  27\penalty0 (9):\penalty0 2522--2535, 2015.

\bibitem[Lines et~al.(2016)Lines, Taylor, and Bagnall]{lines2016hive}
Jason Lines, Sarah Taylor, and Anthony Bagnall.
\newblock Hive-cote: The hierarchical vote collective of transformation-based
  ensembles for time series classification.
\newblock In \emph{2016 IEEE 16th international conference on data mining
  (ICDM)}, pages 1041--1046. IEEE, 2016.

\bibitem[Shifaz et~al.(2020)Shifaz, Pelletier, Petitjean, and
  Webb]{shifaz2020ts}
Ahmed Shifaz, Charlotte Pelletier, Fran{\c{c}}ois Petitjean, and Geoffrey~I
  Webb.
\newblock Ts-chief: A scalable and accurate forest algorithm for time series
  classification.
\newblock \emph{Data Mining and Knowledge Discovery}, pages 1--34, 2020.

\bibitem[Wang et~al.(2017)Wang, Yan, and Oates]{wang2017time}
Zhiguang Wang, Weizhong Yan, and Tim Oates.
\newblock Time series classification from scratch with deep neural networks: A
  strong baseline.
\newblock In \emph{2017 International joint conference on neural networks
  (IJCNN)}, pages 1578--1585. IEEE, 2017.

\bibitem[Karim et~al.(2017)Karim, Majumdar, Darabi, and Chen]{karim2017lstm}
Fazle Karim, Somshubra Majumdar, Houshang Darabi, and Shun Chen.
\newblock Lstm fully convolutional networks for time series classification.
\newblock \emph{IEEE access}, 6:\penalty0 1662--1669, 2017.

\bibitem[Wang et~al.(2015)Wang, Ouyang, Wang, and Lu]{wang2015visual}
Lijun Wang, Wanli Ouyang, Xiaogang Wang, and Huchuan Lu.
\newblock Visual tracking with fully convolutional networks.
\newblock In \emph{Proceedings of the IEEE international conference on computer
  vision}, pages 3119--3127, 2015.

\bibitem[Tachibana et~al.(2018)Tachibana, Uenoyama, and
  Aihara]{tachibana2018efficiently}
Hideyuki Tachibana, Katsuya Uenoyama, and Shunsuke Aihara.
\newblock Efficiently trainable text-to-speech system based on deep
  convolutional networks with guided attention.
\newblock In \emph{2018 IEEE International Conference on Acoustics, Speech and
  Signal Processing (ICASSP)}, pages 4784--4788. IEEE, 2018.

\bibitem[Jung et~al.(2017)Jung, Choi, Jung, Lee, Kwon, and
  Young~Jung]{jung2017resnet}
Heechul Jung, Min-Kook Choi, Jihun Jung, Jin-Hee Lee, Soon Kwon, and Woo
  Young~Jung.
\newblock Resnet-based vehicle classification and localization in traffic
  surveillance systems.
\newblock In \emph{Proceedings of the IEEE conference on computer vision and
  pattern recognition workshops}, pages 61--67, 2017.

\bibitem[Bahdanau et~al.(2016)Bahdanau, Chorowski, Serdyuk, Brakel, and
  Bengio]{bahdanau2016end}
Dzmitry Bahdanau, Jan Chorowski, Dmitriy Serdyuk, Philemon Brakel, and Yoshua
  Bengio.
\newblock End-to-end attention-based large vocabulary speech recognition.
\newblock In \emph{2016 IEEE international conference on acoustics, speech and
  signal processing (ICASSP)}, pages 4945--4949. IEEE, 2016.

\bibitem[Chen et~al.(2019)Chen, Fan, Xu, Yan, Kalantidis, Rohrbach, Yan, and
  Feng]{chen2019drop}
Yunpeng Chen, Haoqi Fan, Bing Xu, Zhicheng Yan, Yannis Kalantidis, Marcus
  Rohrbach, Shuicheng Yan, and Jiashi Feng.
\newblock Drop an octave: Reducing spatial redundancy in convolutional neural
  networks with octave convolution.
\newblock In \emph{Proceedings of the IEEE/CVF International Conference on
  Computer Vision}, pages 3435--3444, 2019.

\bibitem[Karim et~al.(2019)Karim, Majumdar, and Darabi]{karim2019insights}
Fazle Karim, Somshubra Majumdar, and Houshang Darabi.
\newblock Insights into lstm fully convolutional networks for time series
  classification.
\newblock \emph{IEEE Access}, 7:\penalty0 67718--67725, 2019.

\bibitem[Dau et~al.(2019)Dau, Bagnall, Kamgar, Yeh, Zhu, Gharghabi,
  Ratanamahatana, and Keogh]{dau2019ucr}
Hoang~Anh Dau, Anthony Bagnall, Kaveh Kamgar, Chin-Chia~Michael Yeh, Yan Zhu,
  Shaghayegh Gharghabi, Chotirat~Ann Ratanamahatana, and Eamonn Keogh.
\newblock The ucr time series archive.
\newblock \emph{IEEE/CAA Journal of Automatica Sinica}, 6\penalty0
  (6):\penalty0 1293--1305, 2019.

\bibitem[Fawaz et~al.(2019)Fawaz, Forestier, Weber, Idoumghar, and
  Muller]{fawaz2019deep}
Hassan~Ismail Fawaz, Germain Forestier, Jonathan Weber, Lhassane Idoumghar, and
  Pierre-Alain Muller.
\newblock Deep learning for time series classification: a review.
\newblock \emph{Data Mining and Knowledge Discovery}, 33\penalty0 (4):\penalty0
  917--963, 2019.

\bibitem[Dempster et~al.(2020)Dempster, Petitjean, and
  Webb]{dempster2020rocket}
Angus Dempster, Fran{\c{c}}ois Petitjean, and Geoffrey~I Webb.
\newblock Rocket: exceptionally fast and accurate time series classification
  using random convolutional kernels.
\newblock \emph{Data Mining and Knowledge Discovery}, 34\penalty0 (5):\penalty0
  1454--1495, 2020.

\bibitem[Lines et~al.(2018)Lines, Taylor, and Bagnall]{lines2018time}
Jason Lines, Sarah Taylor, and Anthony Bagnall.
\newblock Time series classification with hive-cote: The hierarchical vote
  collective of transformation-based ensembles.
\newblock \emph{ACM Transactions on Knowledge Discovery from Data}, 12\penalty0
  (5), 2018.

\end{thebibliography}






\end{document}